\documentclass[runningheads]{llncs}

\usepackage{eccv}

\usepackage{eccvabbrv}

\usepackage{tabularx}
\usepackage{pifont}
\usepackage{graphicx}
\usepackage{booktabs}

\usepackage[accsupp]{axessibility}  

\usepackage{hyperref}

\usepackage{orcidlink}

\begin{document}

\title{HANS: A Handwritten Answer Sheet Dataset for Noisy Hybrid Document Parsing
} 

\titlerunning{HANS: Noisy Hybrid Document Parsing}


\title{
HANS: A Handwritten Answer Sheet Dataset for
Noisy Hybrid Document Parsing
}

\titlerunning{HANS: Noisy Hybrid Document Parsing}

\author{
Xiazhen Wu \and
Wansong Qin \and
Yangbin Zheng \and
Liangda Fang \and
Zhan Li \and
Xiujie Huang \and
Liushen Zhou  \and
Quanlong Guan
}
\authorrunning{X.~Wu et al.}

\institute{
Jinan University, Guangzhou, China\\
\email{%
wuxiazhen@stu2024.jnu.edu.cn\\
\{songbai,zybbyz\}@stu2025.jnu.edu.cn\\
\{fangld,lizhan,t\_xiujie,crystal,gql\}@jnu.edu.cn
}
}

\maketitle

\begin{abstract}
  Intelligent grading and automated scoring technologies constitute critical infrastructure for smart education. However, existing document parsing and handwriting recognition benchmarks are predominantly designed for well-structured printed documents or isolated mathematical expressions, lacking datasets that capture the complex characteristics inherent to student answer sheets, including multi-line derivation processes, heterogeneous mixtures of text and mathematical formulae, and noise artifacts such as strikethroughs. To address this gap, we introduce HANS, the first dataset explicitly constructed for real-world educational scenarios, encompassing mathematical expressions, natural language text, hand-drawn tables, and diverse noise patterns including corrections and deletions, accompanied by fine-grained annotations that establish a reliable foundation for robust recognition research. Building upon HANS, we propose NA-GOT, an end-to-end framework that achieves two-stage noise suppression through a lightweight noise suppression module operating at the feature level, complemented by a noise-aware attention mechanism incorporated into the decoding stage. Experimental results demonstrate that HANS poses substantial challenges to existing methods, while NA-GOT achieves significant improvements in both accuracy and stability for answer process recognition. The dataset will be made publicly available upon publication.
\end{abstract}

\section{Introduction}
\label{sec:intro}

Intelligent grading and automated scoring technologies are increasingly emerging as pivotal driving forces in the advancement of smart education. In contrast to outcome-oriented evaluation paradigms that focus exclusively on final answers, modern pedagogical practice, particularly in mathematics education, places greater emphasis on understanding and diagnosing students' problem-solving thought processes. This demands that intelligent systems be capable of recognizing the derivation steps and natural language explanations written by students on answer sheets, thereby enabling step-wise scoring, error cause analysis, and personalized feedback. Consequently, the accurate recognition of students' handwritten problem-solving processes constitutes a critical component of intelligent grading systems, as its accuracy and robustness directly determine the viability of downstream scoring and diagnostic tasks.

  Recent years have witnessed substantial progress in document parsing and handwriting recognition. Document parsing methods are predominantly designed for well-structured printed documents, with representative approaches including multi-stage pipeline \cite{1} processing and end-to-end vision-language models \cite{2,3,4,5,6}, exemplified by benchmark datasets such as OmniDocBench \cite{7} and FOX \cite{8}. Concurrently, research on handwriting recognition using classical benchmarks — including IAM \cite{9}, CROHME \cite{10}, and ICDAR Math — has driven an architectural evolution from CNN and RNN-CTC paradigms toward Transformer-based \cite{11} encoders with Seq2Seq decoders and unified vision-language modeling frameworks.

\begin{figure}[tb]
  \centering
  \includegraphics[height=5cm]{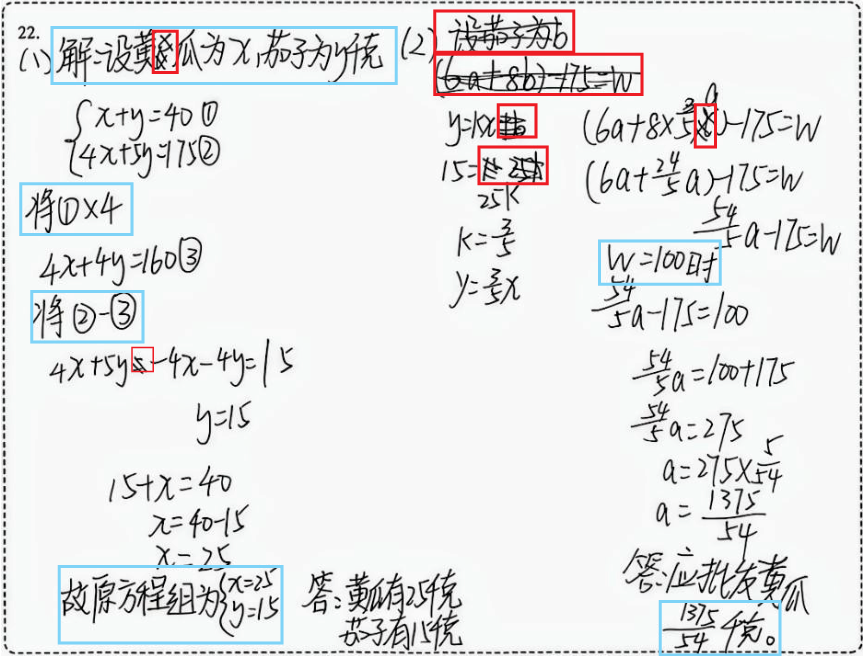}
  \caption{Illustration of representative characteristics of the HANS dataset. HANS is a real-world student handwritten mathematics answer sheet dataset, encompassing the dual challenges of complex formula-text interleaving (blue bounding boxes) and correction noise artifacts (red bounding boxes).
  }
  \label{fig1}
\end{figure}

  Nevertheless, existing benchmarks exhibit several critical limitations. 1) Prevailing datasets predominantly focus on printed academic documents, neglecting real-world handwritten document types such as student answer sheets. 2) There is a systematic absence of research on handwriting noise: most benchmarks assume high document quality and lack systematic annotation and targeted evaluation of common noise artifacts such as strikethroughs, erasures, and overwriting. 3) Existing handwriting datasets are confined to either pure text lines or isolated mathematical expressions; in particular, handwritten mathematical expression recognition has largely been restricted to single-line expressions, with limited investigation into continuous multi-line writing scenarios involving interleaved text and formulae.

  To address these limitations, we introduce HANS, the first dataset explicitly constructed for real-world educational scenarios. HANS is sourced from scanned student answer sheets collected during eighth-grade mathematics quality assessment examinations conducted over three semesters in a municipal school district, specifically from the open-ended mathematical problem-solving sections. To protect student privacy, all personally identifiable information was cropped prior to data release and use. Unlike prior datasets comprising pure text or isolated formulae, the student problem-solving processes in HANS encompass mathematical expressions, natural language text, hand-drawn tables, and manual strikethroughs, faithfully reflecting the typical interplay between mixed handwriting and structured writing characteristic of authentic examination contexts, as shown in \cref{fig1} . More critically, we provide fine-grained annotations across distinct writing modalities within the problem-solving process, including content-level annotations for expressions, text, and tables, as well as spatial annotations for noise elements such as strikethroughs and deletions. These annotations furnish a supervisory foundation for explicitly distinguishing noise from valid handwriting, thereby supporting subsequent robust modeling and analysis.

  Building upon the noise annotations provided by HANS, we further propose NA-GOT, an end-to-end noise-aware recognition framework extending the GOT \cite{12} architecture. The proposed framework suppresses noise interference through a two-stage mechanism. In the first stage, we design a Lightweight Noise Suppression Module (LNSM) that attenuates noise at the feature level. In the second stage, we introduce a noise-aware attention mechanism that reduces attention weights assigned to noise-bearing regions via learned noise bias terms. Experimental results demonstrate that incorporating the noise-aware mechanism yields significant improvements in both recognition accuracy and output stability for answer sheet problem-solving process recognition.

  In summary, the contributions of this paper are as follows:

1. We construct HANS, a real-world educational answer sheet dataset encompassing mixed handwritten problem-solving processes with heterogeneous content types including mathematical expressions, natural language text, and tables. Fine-grained annotations are provided, with particular emphasis on spatial bounding box annotations for real-world noise artifacts such as strikethroughs and deletions, establishing a foundation for robust recognition research.

2. We propose NA-GOT, an end-to-end noise-aware recognition framework that achieves two-stage noise suppression through a Lightweight Noise Suppression Module(LNSM) and a noise-aware attention mechanism.

3. Extensive experimental results demonstrate that the HANS dataset presents substantial challenges to current state-of-the-art methods, and validate that the proposed approach significantly outperforms existing methods.

\section{Related Work}

\subsection{Document Parsing and Handwriting Recognition Datasets}
In the domain of document parsing, datasets such as DocLayNet \cite{13} and OmniDocBench \cite{7} provide important benchmarks for document layout segmentation and multi-task learning, with particular significance for layout element detection and segmentation. Concurrently, datasets including FUNSD \cite{14} and FOX have advanced the parsing of form and handwritten documents, encompassing diverse components such as tables, text, and formulae while extending support to more complex document structures. The data distributions of these benchmarks are predominantly composed of printed text, with relatively limited representation of handwritten content, and systematic investigation of noise-related challenges remains largely absent. Existing handwriting recognition datasets, such as IAM, CROHME, and MathWriting \cite{15} , primarily focus on offline handwriting recognition and mathematical formula transcription tasks. Although substantial progress has been achieved in handwritten text and mathematical expression recognition, these datasets continue to suffer from limited diversity in writing modalities.

\subsection{Document Parsing and Handwriting Recognition Models}
In the area of generative document understanding, models such as Nougat \cite{16} and GOT-OCR2.0 realize end-to-end conversion from document images to structured text via visual Transformer architectures, supporting joint modeling of heterogeneous content types. LayoutLM \cite{17} achieves strong performance through joint modeling of layout and content by integrating visual and textual information within a unified framework. In the field of handwriting recognition, TrOCR \cite{18} adopts an architecture comprising an image Transformer encoder coupled with a text Transformer decoder, enabling efficient end-to-end handwritten text transcription. In the domain of handwritten mathematical expression recognition, works such as TAMER introduce tree-structure-aware modules to jointly optimize sequence prediction and tree structure prediction tasks, thereby improving model generalization over complex mathematical expression structures and demonstrating strong adaptability — particularly for the recognition of intricate formulae and expressions.

Despite the remarkable progress achieved in existing document parsing and handwriting recognition techniques, current methodologies remain inadequate in addressing the noise-corrupted and structurally complex handwriting characteristics inherent to educational answer sheets. Consequently, future research should build upon existing datasets and methods to develop more specialized datasets and more adaptive model frameworks, with the aim of enhancing robustness and accuracy in the educational answer sheet recognition task.

\section{HANS Dataset}
\label{sec:dataset}

\subsection{Dataset Construction and Statistics}
We construct HANS, a dataset designed for real-world educational answer sheet scenarios. The data are sourced from eighth-grade mathematics quality assessment examinations conducted over three semesters in a municipal middle school district in China, specifically from the open-ended problem-solving sections comprising six mathematical questions per examination, drawn from scanned student answer sheets, as shown in \cref{tab1}. The dataset contains 5,213 answer question images, covering 745 students across 18 distinct problem types. To protect student privacy, all personally identifiable information was cropped prior to data release and use.

\begin{table}[tb]
  \caption{Comparison of HANS against existing datasets.HS denotes the number of handwritten samples,and Level indicates the sample granularity(page or line). HME, HT, and NA indicate the presence of handwritten mathematical expressions,handwritten text, and noise annotations,respectively.
  }
  \label{tab1}
  \centering
  \begin{tabular}{@{}llllll@{}}
    \toprule
    Dataset\hspace{8mm} & HS\hspace{8mm} & Level\hspace{8mm} & HME\hspace{8mm} & HT\hspace{8mm} & NA \\
    \midrule
    OmniDocBench & 116 & Pages & \ding{51} & \ding{51} & \ding{55}\\
    CC-OCR \cite{19} & 200 & Pages & \ding{51} & \ding{55} & \ding{55}\\
    FOX & 0 & Pages & \ding{55} & \ding{55} & \ding{55}\\
    IAM & 1539 & Lines & \ding{55} & \ding{51} & \ding{55}\\
    CROHME & 8836 & Lines & \ding{51} & \ding{55} & \ding{55}\\
    \textbf{HANS(ours)} & \textbf{5213} & \textbf{Pages} & \textbf{\ding{51}} & \textbf{\ding{51}} & \textbf{\ding{51}}\\
  \bottomrule
  \end{tabular}
\end{table}

  The writing patterns in HANS exhibit considerable complexity. Within a single page, natural language explanations and mathematical expressions frequently appear in interleaved configurations, accompanied by extensive multi-line derivation steps, chains of equations, and cross-line intermediate results that collectively encode rich procedural information. The mathematical expressions themselves present high structural complexity, commonly featuring fractions, radicals, superscripts and subscripts, nested parentheses, and diverse geometric and logical symbols, further compounding the difficulty of symbol-level accurate transcription and structural consistency constraints. In addition, HANS contains a non-trivial proportion of hand-drawn tables, whose cells typically contain both natural language and mathematical content simultaneously, substantially increasing the coupling between structural layout recognition and content transcription. Furthermore, the answer sheets exhibit pervasive real-world noise and correction behaviors characteristic of authentic examination contexts, including erasures, strikethroughs, and overwriting, which introduce stroke discontinuities, stroke superposition, local occlusion, and structural degradation. These artifacts result in severe spatial entanglement between valid handwriting and noise traces, significantly elevating the uncertainty and difficulty faced by recognition models.

\begin{figure}[tb]
  \centering
  \includegraphics[height=5cm]{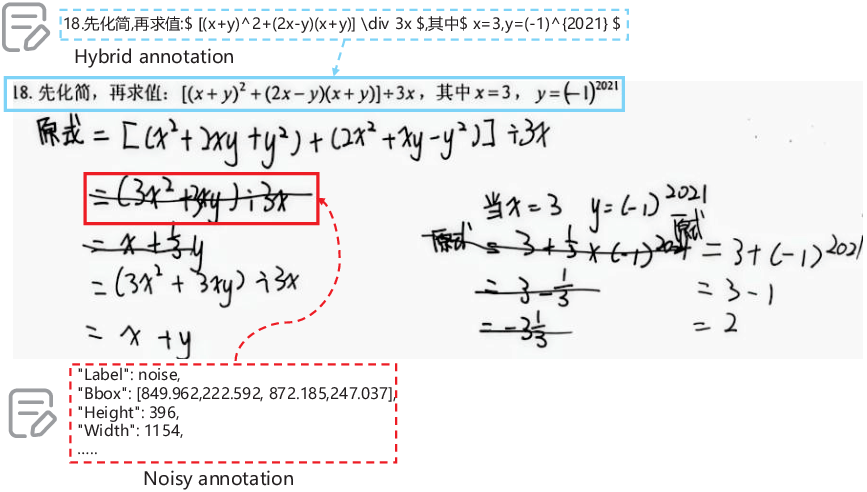}
  \caption{Characteristics of the HANS annotated dataset. Each sample in the dataset is annotated with both the complete problem-solving process transcription and the corresponding correction noise regions. Blue bounding boxes denote the annotation of interleaved mathematical expressions and natural language text, while red bounding boxes indicate the spatial annotations of correction noise artifacts.
  }
  \label{fig2}
\end{figure}

\subsection{Annotation Details}
Unlike benchmarks that provide only final answer transcriptions or annotations for isolated formulae, we provide annotations more closely aligned with the requirements of automated grading. The fundamental sample unit of the dataset is defined as the answer region image corresponding to each open-ended problem, with each sample containing two categories of annotations: (1) problem-solving process recognition annotations, and (2) non-semantic noise region annotations for erasures and strikethroughs. Full details are illustrated in \cref{fig2}.

\subsubsection{Problem-Solving Process Recognition Annotations}
Each sample is provided with a recognition transcription corresponding to the complete problem-solving process visible in the image; content that has been deleted by the student is excluded from the final transcription. In terms of structural representation, the problem-solving process is segmented according to the semantic information of each handwritten line, strictly preserving the original line breaks as written. The logical ordering between lines is inferred from handwriting layout conventions, with line break symbols employed to denote logical sequencing.

\subsubsection{Noise Region Detection Annotations}
For images containing noise, we provide pixel-level annotation files for noise regions, stored in JSON format. All noise instances within an image are uniformly assigned the label noise and are permitted to adopt one of two geometric annotation structures: block-level erasures or strikethroughs are annotated using rectangle, defined by two diagonal corner coordinates, while irregularly shaped corrections or localized ink clusters are annotated using polygon, defined by a closed multi-point contour. In cases where noise regions partially overlap with valid content, conservative boundary fitting is applied to minimize erroneous coverage of effective derivation areas.
For strikethrough scenarios, annotations primarily delineate the extent of visible occlusion caused by deletion strokes. For overwriting scenarios, annotations encompass both the discarded content and the superimposed strokes, while endeavoring to preserve the final legible and correct writing. For erasure scenarios, annotations cover both erasure residuals and regions rendered illegible by the erasing action.

\subsection{Annotation Pipeline}
To ensure both annotation quality and operational efficiency, we design a three-stage pipeline for the HANS annotation task, comprising automatic pre-annotation, annotator correction, and expert quality inspection.

  \textbf{Automatic pre-annotation}. We employ state-of-the-art detection and recognition models to generate pre-annotations for both noise region detection and problem-solving process transcription. Specifically, YOLOv8 \cite{20} is adopted for the automatic localization of student correction regions, while DeepSeek-OCR \cite{21} is employed for preliminary content recognition of answer sheets, encompassing annotations for text, mathematical expressions, and tables.

  \textbf{Annotator correction}. Following automatic pre-annotation, human annotators systematically revise the initial annotation results. For student correction regions, annotators identify and supplement any missed noise areas, and refine the bounding box boundaries to more precisely conform to the extents of the correction artifacts. Each character is individually verified to ensure transcription accuracy. With respect to the overall problem-solving process, annotators perform character-level inspection and correction of the preliminary recognition drafts, ensuring the accurate representation of text, mathematical symbols, and their structural relationships. Omitted content is supplemented where necessary, and output formatting is standardized to satisfy the requirements of subsequent model training and evaluation.

  \textbf{Quality inspection}. Despite the comprehensive corrections performed by annotators, the structural complexity of mathematical formulae and tables may nonetheless introduce residual errors. Accordingly, domain experts conduct a dedicated quality inspection pass to verify annotation accuracy and ensure consistency across the dataset.

\section{Method}
\label{sec:method}
\begin{figure}[tb]
  \centering
  \includegraphics[height=5cm]{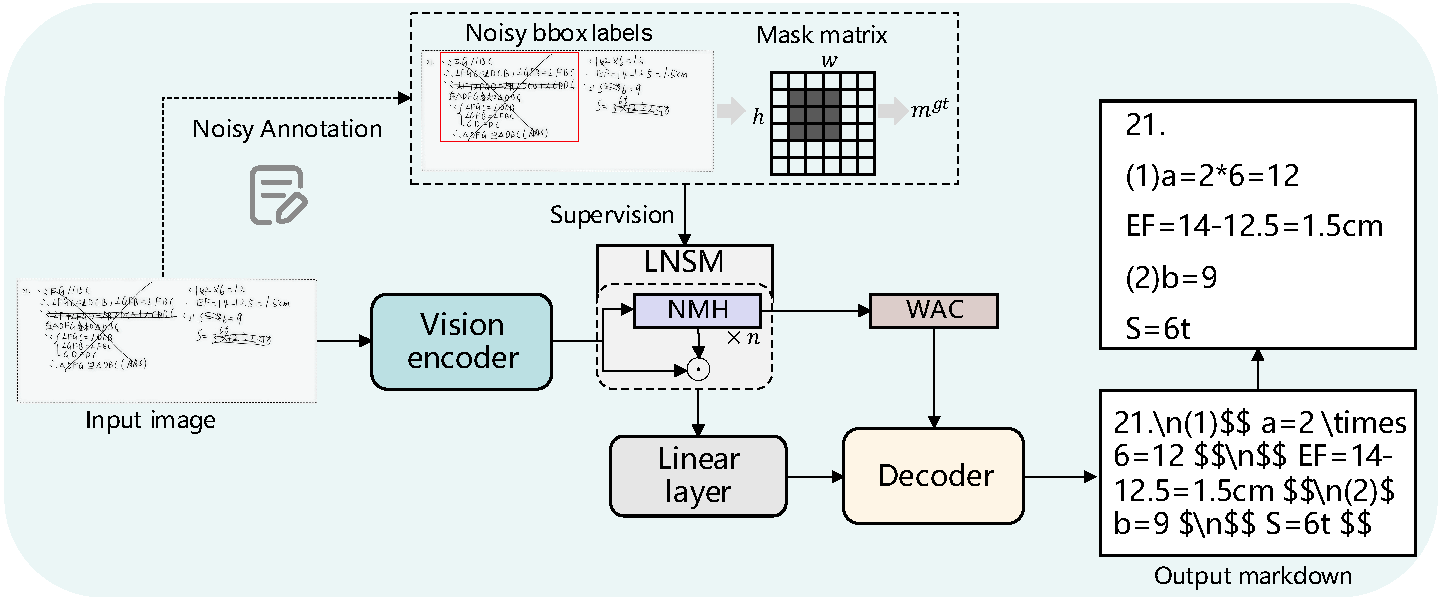}
  \caption{Overall architecture of the NA-GOT framework.
  }
  \label{fig3}
\end{figure}

We propose NA-GOT, a noise-aware end-to-end answer sheet recognition framework, as illustrated in \cref{fig3}, designed to mitigate the interference of real-world noise artifacts, including erasures, strikethroughs, and overwriting on the accurate recognition of valid handwritten characters. Building upon GOT, the proposed framework suppresses noise through a two-stage mechanism. In the first stage, we design a Lightweight Noise Suppression Module (LNSM) that attenuates noise influence at the feature level. In the second stage, we introduce a Noise-aware attention mechanism that reduces attention weights assigned to noise-bearing regions via learned noise bias terms. Specifically, the input image is first processed by a visual encoder to extract patch-level features, while a mask matrix generated from noise annotations provides noise-aware supervision signals during training. The noise-aware module performs weighted suppression of features according to the predicted noise probabilities, and subsequently modulates the decoder's attention distribution to direct the model's focus toward valid handwriting regions. The resulting visual features are then projected via a linear layer and fed into the decoder, which generates the final Markdown-formatted output, achieving complete transcription of natural language text, mathematical formulae, and step-by-step derivations.

\subsection{Lightweight Noise Suppression Module}
We first construct patch-level supervision signals aligned with the visual patch tokens from the noise annotations. Specifically, for each image, all annotations are first converted into a binary pixel mask matrix $M$ according to the original image resolution. To maintain consistency with the patch token layout of the GOT visual encoder, $M$ is uniformly resized to a resolution of 1024×1024, followed by 16×16 average pooling to compute the noise occupancy ratio over the patch grid. The resulting grid is subsequently flattened into patch-level noise labels $m^gt$ in one-to-one correspondence with the visual tokens, where each entry in $m^gt$ represents the proportion of noise pixels within the corresponding patch.

We takes the patch-level noise labels $m^gt$ and the output features $V$ of the visual encoder as inputs to the Noise Mask Head (NMH), which determines whether each patch constitutes a noise patch. Concretely, NMH employs a two-layer MLP to predict a noise confidence score $p$ for each patch token:

$$p = \sigma(\text{MLP}(V))$$

This design enables the model to adaptively predict noise distributions during inference without relying on manual annotations or additional noise inputs, while simultaneously providing stable noise priors for subsequent token gating and the decoder's noise-aware attention bias, thereby enhancing the robustness of downstream sequence generation.

Leveraging the obtained noise confidence scores $p$, we further suppress the representational magnitude of noise tokens at the visual token level. Specifically, the noise confidence scores p are mapped to gating coefficients, which are applied to scale the visual tokens on a per-token basis:

$$\tilde{V} = V \odot \bigl( 1 - \alpha \cdot p \bigr)$$

where $\odot$ denotes element-wise multiplication, and $\alpha$ is a hyperparameter controlling the gating strength.

\subsection{Weighted Attention Correction}
While LNSM modulates the representational magnitude of visual tokens, the decoder may nevertheless assign substantial attention weights to noise tokens. Since NA-GOT concatenates visual patch tokens and text tokens into a unified sequence, cross-modal alignment occurs within the decoder. To suppress the interference of noise visual tokens on the generation process, Weighted Attention Correction (WAC) introduces a negative bias derived from the noise probabilities p output by NMH into the decoder's self-attention layers:

$$b = -\beta \cdot p$$

where $b$ denotes the bias term and $\beta$ is a hyperparameter controlling the penalty strength. A higher noise probability yields a larger bias, resulting in a lower attention weight for the corresponding patch following the softmax operation.

This mechanism penalizes noise patches during attention computation, directing the decoder's focus toward valid handwriting regions, thereby improving both the accuracy and structural integrity of the generated Markdown output in complex handwritten answer sheet scenarios.

\subsection{Loss Function}
The NMH predicts a noise confidence score p for each visual patch token, and the noise labels $m^{gt}$ are constructed from the noise annotations. A binary cross-entropy loss with positive sample weighting is applied to compute the noise mask loss $L_{mask}$. Given the input image and the corresponding prompt, the model autoregressively generates the target sequence, with cross-entropy loss computed as the primary recognition loss $L_{ocr}$. The final training objective is defined as:

$$\mathcal{L} = \mathcal{L}_{\text{ocr}} + \lambda_{\text{mask}} \, \mathcal{L}_{\text{mask}}$$

where $\lambda_{\text{mask}}$ denotes the noise branch weight, controlling the influence of noise supervision on the primary recognition task.

\section{Experiments}

\subsection{Experimental Setup}

\subsubsection{Implementation Details}
We employ the proposed NA-GOT model for both training and evaluation. For the LNSM, patch-level noise supervision signals are generated from noise annotations during training to supervise the NMH in predicting noise distributions. The key hyperparameters adopted are $\alpha$ = 0.7 and $\beta$ = 2.0.

\subsubsection{Dataset and Evaluation Metrics} 
We conduct all experiments on the proposed HANS dataset to evaluate whether the proposed method effectively addresses the robustness challenges inherent to answer sheet recognition. The dataset is partitioned into training and test sets at an 8:2 ratio, with samples containing noise regions comprising three-quarters of the training set. Evaluation is performed using sequence alignment metrics commonly adopted in document parsing research, including BLEU, Edit Distance, METEOR, Precision, Recall, and F1-score.

\subsubsection{Baselines} To comprehensively evaluate the proposed method, we select three categories of representative baselines for comparison: (1) General VLMs: Qwen3-VL \cite{22} and InternVL3.5 \cite{23}; (2) Specialized VLMs: DeepSeek-OCR2, PaddleOCR-VL1.5 \cite{24}, MinerU2.7.6-VLM \cite{25}, and GOT; (3) Pipeline Tools: PaddleOCR PP-StructureV3 \cite{26} and MinerU2.7.6-Pipeline. To provide a rigorous comparison, GOT is additionally fine-tuned on the full HANS training set prior to evaluation, thereby demonstrating the superiority of the proposed approach.

\subsection{Comparison to Previous Approaches}
We present a comprehensive comparison of NA-GOT against prior methods on the HANS dataset in \cref{tab2}, with all baselines organized into three groups by model category. NA-GOT consistently surpasses all baselines across General VLMs, Specialized VLMs, and Pipeline Tools. Pipeline Tools exhibit notably inferior performance on this task, while General VLMs and Specialized VLMs achieve moderate performance. Among these, Qwen3-VL delivers the strongest results, indicating that powerful general-purpose VLMs possess considerable document understanding and generation capabilities, yet remain susceptible to noise interference in the challenging noisy answer sheet scenario. GOT transferred directly to this dataset yields relatively low performance, suggesting a substantial domain shift between this task distribution and the original training data — particularly with respect to noise patterns and the heterogeneous output requirements imposed by interleaved formula-text handwriting. This further underscores the value of HANS in filling a critical data gap in the document parsing domain. In contrast, GOT fine-tuned on HANS achieves significant performance gains, validating the effectiveness of our annotations in enabling the model to rapidly adapt to the target data distribution. Compared to the fine-tuned GOT baseline, our proposed method achieves consistent improvements across all evaluation metrics, with particularly pronounced gains on heavily noisy samples. These results demonstrate that the proposed Lightweight Noise Suppression Module(LNSM) effectively learns to extract noise-invariant features, while the Noise-aware attention mechanism penalizes noise-bearing positions at the attention layer, thereby mitigating noise-induced misalignment and repetitive generation.

\begin{table}[tb]
  \caption{Experimental results of NA-GOT and baseline models on the HANS dataset.
  }
  \label{tab2}
  \centering
  \begin{tabular}{@{}lllllll@{}}
    \toprule
    Method & BLEU$\uparrow$ & Edit Distance$\downarrow$ & METEOR$\uparrow$ & Precision$\uparrow$ & Recall$\uparrow$ & F1-score$\uparrow$\\
    \midrule
    PaddleOCR PP-V3 & 0.177 & 0.659 & 0.336 & 0.624 & 0.546 & 0.566\\
    MinerU2.7.6-pipeline & 0.064 & 0.734 & 0.218 & 0.482 & 0.431 & 0.439\\
    \midrule
    Qwen3-VL & 0.433 & 0.445 & 0.586 & 0.734 & 0.731 & 0.724\\
    InternVL3.5 & 0.196 & 0.650 & 0.364 & 0.660 & 0.554 & 0.650\\
    \midrule
    Paddle-VL1.5 & 0.349 & 0.526 & 0.497 & 0.722 & 0.665 & 0.678\\
    MinerU2.7.6-VLM & 0.274 & 0.584 & 0.485 & 0.611 & 0.677 & 0.629\\
    Deepseek-OCR2 & 0.338 & 0.533 & 0.505 & 0.655 & 0.665 & 0.649\\
    \midrule
    GOT & 0.098 & 0.789 & 0.231 & 0.426 & 0.470 & 0.409\\
    GOT(fine tuned) & 0.577 & 0.324 & 0.687 & 0.812 & 0.806 & 0.801\\
    \textbf{NA-GOT(fine tuned)} & \textbf{0.594} & \textbf{0.306} & \textbf{0.701} & \textbf{0.821} & \textbf{0.813} & \textbf{0.810}\\
  \bottomrule
  \end{tabular}
\end{table}

\subsection{Ablation Study}

We conduct an ablation study by removing individual noise-aware components from the full NA-GOT model to evaluate their respective contributions, as reported in \cref{tab3}. The w/o WAC variant achieves a BLEU score of 0.589 and an Edit Distance of 0.314, indicating that the noise prediction and token gating mechanism alone can effectively suppress noise interference at the feature level. The w/o token gating variant yields lower BLEU and F1-score than the w/o WAC variant, suggesting that feature-level token gating contributes more substantially to robust recognition than decoder-level attention correction alone. Nevertheless, both ablated variants underperform the full NA-GOT model. The full model achieves the best overall performance across all evaluation metrics, demonstrating that token gating and WAC provide complementary benefits at the representation and decoding stages, respectively.

\begin{table}[tb]
  \caption{Ablation study of token gating and Weighted Attention Correction (WAC) in NA-GOT.
  }
  \label{tab3}
  \centering
  \begin{tabular}{@{}lllllll@{}}
    \toprule
    Method & BLEU$\uparrow$ \hspace{1mm} & Edit Distance$\downarrow$ \hspace{1mm} & METEOR$\uparrow$ \hspace{1mm} & Precision$\uparrow$ \hspace{1mm} & Recall$\uparrow$ \hspace{1mm} & F1-score$\uparrow$\\
    \midrule
     \textbf{NA-GOT} & \textbf{0.594} & \textbf{0.306} & \textbf{0.701} & \textbf{0.821} & \textbf{0.813} & \textbf{0.810}\\
     w/o WAC  & 0.589 & 0.314 & 0.698 & 0.820 & 0.809 & 0.808\\
    w/o Token Gating & 0.580 & 0.321 & 0.691 & 0.815 & 0.808 & 0.803\\
  \bottomrule
  \end{tabular}
\end{table}

\subsection{Hyperparameter experimentation}

Furthermore, we conduct ablation experiments on the two key hyperparameters, $\alpha$ and $\beta$. As reported in \cref{tab4}, the gating strength coefficient $\alpha$ is found to exert a significant influence on model performance. When $\alpha$ = 0.5, noise suppression is insufficient, allowing noise tokens to propagate into the decoder with considerable magnitude, resulting in marginally lower BLEU and F1-score. When $\alpha$ = 0.9, excessive suppression impairs a portion of valid features, leading to slight degradation in BLEU and METEOR. The model achieves optimal overall performance at $\alpha$ = 0.7, indicating that moderate representation-level noise suppression effectively reduces noise-induced erroneous generation without sacrificing coverage of critical information.

The hyperparameter $\beta$ governs the penalty strength imposed on noise patches at the attention layer, with results reported in \cref{tab5}. When $\beta$ = 1.0, the penalty is insufficient, causing the model to produce erroneous outputs in certain noise-bearing regions. When $\beta$ = 3.0, the excessively large penalty introduces mild over-suppression, marginally weakening the decoder's attention allocation toward valid patches. At $\beta$ = 2.0, the model exhibits a stronger tendency to aggregate information from non-noise regions, yielding optimal performance across all evaluation metrics.

\begin{table}[tb]
  \caption{Analysis of the influence of hyperparameter $\alpha$ on model performance.
  }
  \label{tab4}
  \centering
  \begin{tabular}{@{}lllllll@{}}
    \toprule
    $\alpha$ & BLEU$\uparrow$ \hspace{1mm} & Edit Distance$\downarrow$ \hspace{1mm} & METEOR$\uparrow$ \hspace{1mm} & Precision$\uparrow$ \hspace{1mm} & Recall$\uparrow$ \hspace{1mm} & F1-score$\uparrow$\\
    \midrule
    0.5  & 0.592 & 0.307 & 0.698 & 0.818 & 0.806 & 0.807\\
    \textbf{0.7} & \textbf{0.594} & \textbf{0.306} & \textbf{0.701} & \textbf{0.821} & \textbf{0.813} & \textbf{0.810}\\
    0.9  & 0.593 & 0.306 & 0.699 & 0.820 & 0.810 & 0.809\\
  \bottomrule
  \end{tabular}
\end{table}

\begin{table}[tb]
  \caption{Analysis of the influence of hyperparameter $\beta$ on model performance.
  }
  \label{tab5}
  \centering
  \begin{tabular}{@{}lllllll@{}}
    \toprule
    $\beta$ & BLEU$\uparrow$ \hspace{1mm} & Edit Distance$\downarrow$ \hspace{1mm} & METEOR$\uparrow$ \hspace{1mm} & Precision$\uparrow$ \hspace{1mm} & Recall$\uparrow$ \hspace{1mm} & F1-score$\uparrow$\\
    \midrule
    1.0 & 0.593 & 0.307 & 0.700 & 0.819 & 0.811 & 0.810\\
    \textbf{2.0} & \textbf{0.594} & \textbf{0.306} & \textbf{0.701} & \textbf{0.821} & \textbf{0.813} & \textbf{0.810}\\
    3.0  & 0.591 & 0.308 & 0.697 & 0.817 & 0.807 & 0.808\\
  \bottomrule
  \end{tabular}
\end{table}

\subsection{Visual Analysis}

The recognition results from four models—NA-GOT, Qwen3-VL, DeepSeek-OCR2, and MinerU2.7.6-pipeline—are compared in Figure 4. NA-GOT demonstrates stronger robustness and generation stability in noisy areas, making fewer errors in misinterpreting erasure textures or strikethrough lines as characters or symbols. It maintains a reasonable line-level structure and derivation order.Additionally, NA-GOT shows a higher fidelity to structured formulas, with fractions, parentheses, and equality chains more consistent, and key variables and operators more coherent, thus reducing semantic bias caused by omissions, mistakes, or structural damage.

\begin{figure}[tb]
  \centering
  \includegraphics[height=5cm]{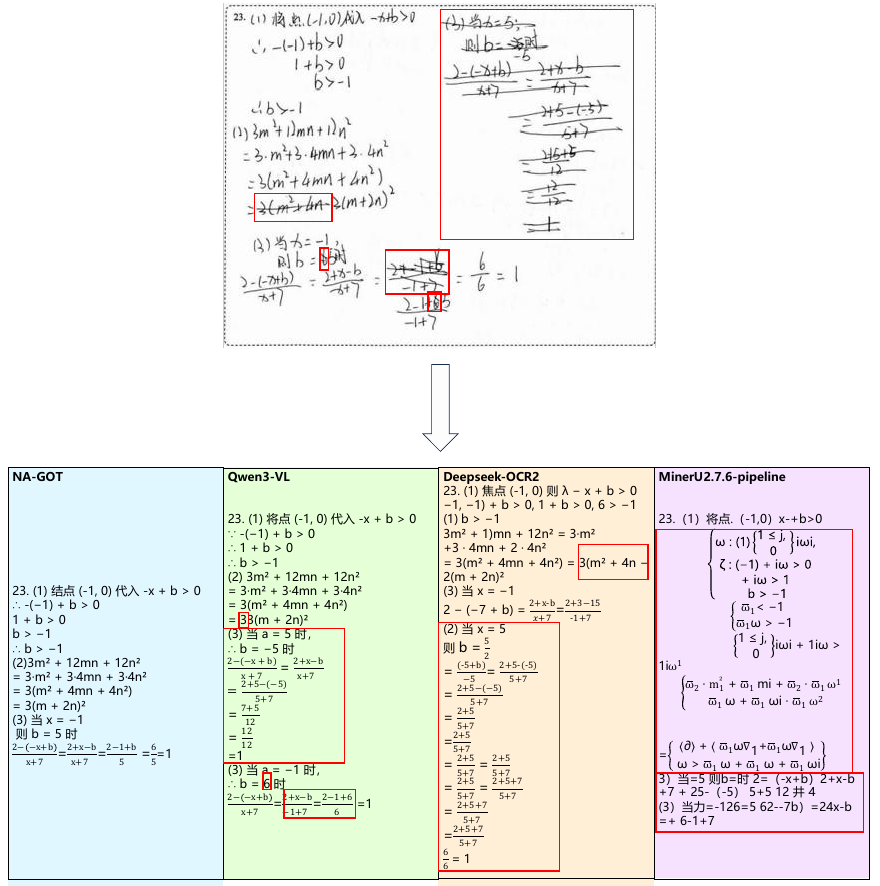}
  \caption{Visualization Results.Figure 4 presents a sample of answer sheet image recognition. The upper part shows the original answer sheet image, with the red box indicating the noise areas due to student corrections. The lower part displays the
recognition results from four models—NA-GOT, Qwen3-VL, DeepSeek-OCR2, and MinerU2.7.6-pipeline—with red boxes highlighting the recognition errors caused by
noise regions in the image
  }
  \label{fig4}
\end{figure}

In contrast, general multimodal models like Qwen3-VL, although generally readable, are more prone to misreading similar-looking variables and numbers near noise boundaries, and tend to experience local misalignment or omission in complex derivations. Specialized VLMs, such as DeepSeek-OCR2, are more prone to decoding instability or structural misreadings under strong noise conditions. For instance, in the red box, DeepSeek-OCR2 shows noticeable repetitive outputs, rendering the entire derivation unusable. Pipeline methods, such as MinerU2.7.6-pipeline, are more likely to miss content or make incorrect identifications, leading to missing steps, disjointed content, or even unrelated structures being introduced.

Overall, this visualization case demonstrates that NA-GOT effectively mitigates the misattention and misalignment issues induced by noise, significantly suppressing the degradation caused by repetitive generation in long sequences. As a result, it provides more stable, structured, and accurate recognition outputs, which are closer to being usable for subsequent automatic grading and diagnosis in real answer sheet scenarios.

\section{Conclusion}
In this paper, we address the problem of recognizing students' handwritten problem-solving processes on answer sheets in smart education scenarios. We introduce HANS, the first dataset specifically constructed for this task, encompassing mathematical expressions, natural language text, hand-drawn tables, and diverse noise artifacts including strikethroughs and deletions, with fine-grained annotations that provide a reliable foundation for robust recognition research. Furthermore, we propose NA-GOT, an end-to-end noise-aware recognition framework that achieves two-stage noise suppression through a lightweight noise suppression module and a noise-aware attention mechanism. Experimental results demonstrate that the HANS dataset presents substantial challenges to existing methods, while NA-GOT achieves significant improvements in both recognition accuracy and output stability for problem-solving process transcription, offering an effective solution for handwritten document parsing in real-world educational scenarios. 



%
%
\bibliographystyle{splncs04}
\bibliography{NA-GOT-1}
\end{document}